\pdfoutput=1
\documentclass[11pt,a4paper]{article}
\usepackage[hyperref]{emnlp2020}
\usepackage{times}
\usepackage{latexsym}

\usepackage{microtype}

\usepackage{graphicx}
\usepackage{verbatim}
\usepackage{siunitx}
\usepackage{enumitem}
\usepackage{subcaption}
\usepackage{bigfoot}
\usepackage{multirow}
\usepackage{hyperref}
\usepackage{tabularx}

\aclfinalcopy % Uncomment this line for the final submission
\title{Bridging Information-Seeking Human Gaze and Machine Reading Comprehension}

\author{Jonathan Malmaud \\
  MIT BCS \\
  \texttt{malmaud@mit.edu} \\\And
  Roger Levy \\
  MIT BCS\\
  \texttt{rplevy@mit.edu} \\\And
  Yevgeni Berzak \\
  MIT BCS\\
  \texttt{berzak@mit.edu} \\
  }

\date{}

\begin{document}
\maketitle
\begin{abstract}
  In this work, we analyze how human gaze during reading comprehension is conditioned on the given reading comprehension question, and whether this signal can be beneficial for machine reading comprehension. To this end, we collect a new eye-tracking dataset with a large number of participants engaging in a multiple choice reading comprehension task. Our analysis of this data reveals increased fixation times over parts of the text that are most relevant for answering the question. Motivated by this finding, we propose making automated reading comprehension more human-like by mimicking human information-seeking reading behavior during reading comprehension. We demonstrate that this approach leads to performance gains on multiple choice question answering in English for a state-of-the-art reading comprehension model.
\end{abstract}

\section{Introduction}

Much of the work in NLP strives to develop systems that are able to perform linguistic tasks similarly to humans. To achieve this goal, one typically provides NLP systems with \emph{human knowledge} about the task at hand. This knowledge can come in the form of linguistic annotations, hand-crafted rules and access to linguistic databases, as well as various model design choices.

In this work, we study the possibility of providing the model with an inductive bias by using \emph{human behavioral signals} based on eye movements in reading as an additional source of information which can guide NLP models to adequately process linguistic input and solve linguistic tasks. As a case study, we examine reading comprehension, a task of central importance for probing both human and machine understanding of text.  
To enable this study, we collect eye movement data from 269 participants who engage in a reading comprehension task using the materials of OneStopQA \cite{berzak2020starc}.

We argue that reading comprehension is a particularly well-suited task for linking human eye movement information to NLP modelling due to the substantial correspondence between reading times and the relevance of the text segment for answering the question. Hahn and Keller \shortcite{hahn2018} have shown this correspondence by establishing increased reading times on the correct answer in a question answering task where answers are named entities. Our study generalizes this result to an arbitrary QA setting, and demonstrates longer reading times for portions of the text which are most pertinent for answering the question correctly.

Building on this observation, we develop a new approach to machine reading comprehension in which the model is directed to \emph{mimic human fixation times} over the text, given the question. The idea behind this approach is to encourage the model to focus on question-relevant information. Specifically, we introduce a multi-task reading comprehension architecture in which a state-of-the-art transformer model jointly performs  question-answering and prediction of the human reading time distribution over the text. 

Our modelling framework is \emph{behavioral}, treating the reading comprehension model itself as a black-box. This leads to both theoretical and practical advantages. From a theoretical perspective, this approach is appealing as it creates a direct parallel to human reading, in which eye movements are an external behavior. Practically, our approach has the advantage of being modular, allowing swapping our model with other reading comprehension models, and the task with other NLP tasks. 

Our experiments demonstrate that our approach leads to consistent gains in question-answering performance across different training regimes, model variants, and on both in- and out-of-domain evaluations. In particular, our model outperforms baseline models with gaze from human reading without exposure to the question. It also performs better than using manual annotations of the textual span critical for answering the question. 

To summarize, we present three contributions:
\begin{enumerate}
    \setlength\itemsep{0em}
    \item We collect an eye-tracking dataset with a large number of participants engaging in free-form multiple choice question answering.
    \item We show that human gaze behavior during question answering is strongly task-conditioned.
    \item We demonstrate that human gaze can improve the performance of a state-of-the-art reading comprehension model.
\end{enumerate}
While this work is a proof of concept and uses a relatively costly data collection procedure, as eye-tracking technology continues to become more ubiquitous and affordable, it will be feasible to perform large scale data collection and deployment of similar approaches for QA and other NLP tasks.

\section{Related Work}

Our work contributes to two areas of research. The first is how human gaze is conditioned on the reading task. This question was previously investigated in the domain of question answering by Hahn and Keller \shortcite{hahn2018}, who collected eye-tracking data in an experimental setup similar to ours for materials from the CNN and Daily Mail corpus \cite{hermann2015}. They demonstrate that reading times on the named entity which is the correct answer to the question are longer if participants are shown the question before reading the passage as compared to ordinary reading. Our work builds on this result, introducing a more general QA setup which is not restricted to questions whose answer is a named entity. Crucially, we further leverage this information for improving machine question answering.

The second research area to which or work contributes is augmenting NLP models with gaze data. In this area, gaze during reading has been used for tasks such as syntactic annotation \cite{barrett2015pos,barrett2015functions,barrett2016,strzyz2019}, text compression \cite{klerke2016}, text readability \cite{gonzalez2017}, Named Entity Recognition \cite{hollenstein2019-entity}, and sentiment classification \cite{mishra2016,mishra2017,mishra2018}. Work on the first four tasks used task-independent eye-tracking corpora, primarily the Dundee corpus \cite{dundee} and GECO \cite{geco2017}. For the task of sentiment classification, the authors used task specific eye-tracking corpora in which the participants were asked to perform sentiment classification. 

Our study differs from this literature in several aspects. First, we address the previously unexplored task of reading comprehension, which has established theoretical and empirical connections to eye movements in reading \citep[][{\it among others}]{just1980,reichle2010eye,rayner2016,hahn2018}.
Also differently from these studies, we cover and directly compare both a task specific reading condition (Hunting) and a task-independent condition (Gathering), as well as both external (Dundee) and corpus specific (OneStopQA) eye-tracking data.

Our QA task can be viewed as a generalization of the work in Mirsha et al. \shortcite{mishra2016,mishra2017,mishra2018}, where instead of being asked about the sentiment of a paragraph, subjects are presented with arbitrary questions.
Our multitask approach for jointly performing the QA task and predicting gaze is similar to Klerke et al. \shortcite{klerke2016}, Berrett et al. \shortcite{barrett2018} and Mishra et al. \shortcite{mishra2018}. In particular, in Equation \ref{loss_equation} we use the same loss term as Barrett et al. \shortcite{barrett2018} which consists of a linear combination of an NLP task loss and gaze prediction loss. Our approach differs from Barrett et al. \shortcite{barrett2018} in that their model uses the gaze predictions as input attention weights for the NLP task, while our model treats gaze only as an output. Our approach provides a parallel to human reading, in which eye movements are an external behavior rather than an input to language processing tasks. Our work differs from Mishra et al. \shortcite{mishra2018} in the model and the use of a single auxiliary objective based on gaze. Finally, we note that in Vajjala et al. \shortcite{vajjala2016} eye-tracking data from ESL learners was collected for 4 articles from the same source of OneStopEnglish articles \cite{vajjala2018onestop} used here, and utilized to study the influence of text difficulty level on fixation measures and reading comprehension. Our work focuses on a different task and a different population of readers.

A large body of work exists on QA, including span prediction (e.g. BiDAF \cite{bidaf2017}), cloze (e.g. \cite{hermann2015}), and multiple choice QA (e.g. Stanford Attentive Reader \cite{chen2016}). Here, we focus on multiple choice QA due to its prevalence in human evaluations of reading comprehension, and use RoBERTa due to its state-of-the-art performance on this task. Further, neural models for QA deploy various notions of \emph{internal attention}. The study of NLP model internal attention has drawn much interest in recent years \citep[][{\it among others}]{adi2016fine,clark2019,serrano2019,kovaleva2019,hoover2019exbert}. In this work we abstract away from model internal dynamics due to their complexity, and the theoretical justification for treating gaze as an external behavior rather than an internal model property. Examination of internal model attention and its relation to human gaze is however an intriguing research direction that we intend to pursue in future work.

\section{Data}

\subsection{Reading Comprehension Data}
\label{rc_data}

We use two reading comprehension resources, OneStopQA \cite{berzak2020starc} and RACE \cite{lai2017race}. \\
\textbf{OneStopQA} is a reading comprehension dataset containing paragraph-level multiple choice reading comprehension questions for 30 Guardian articles (162 paragraphs) taken from the OneStopEnglish dataset \cite{vajjala2018onestop}. Each article is available in three parallel text difficulty levels: the original Advanced text and two simplified versions, Intermediate and Elementary. Each paragraph has three multiple choice reading comprehension questions. All the questions are answerable based on any of the text level versions of the paragraph. We use the Advanced and Elementary text versions, corresponding to 972 question--paragraph pairs.

The answers for each OneStopQA question are structured as follows. \\
\textbf{A} is the correct answer. Answering a question correctly requires information from a textual span in the paragraph called the \emph{critical span}. Importantly, the critical span does not contain the answer in verbatim form.\\
\textbf{B} is a distractor which represents a plausible miscomprehension of the critical span. \\
\textbf{C} is a distractor which is anchored in an additional span in the paragraph, called the distractor span. \\
\textbf{D} is a distractor which has no support in the text.
Both the critical span and the distractor span are annotated manually in the text. 

\textbf{RACE} is the standard dataset in NLP for training and evaluation of multiple choice reading comprehension. It comprises reading comprehension examination materials for middle school and high school students in China. Similarly to OneStopQA, RACE questions are multiple choice, with four possible answers for each question. As opposed to OneStopQA, the questions are based on an entire article rather than a specific paragraph and the answers have no systematic structure with respect to the text. Although RACE has been widely used in NLP, it was recently shown that it has substantial quality assurance drawbacks; 47\% of its questions are guessable by RoBERTa without the passage, and 18\% do not have a unique correct answer \cite{berzak2020starc}. We therefore treat RACE as a secondary evaluation benchmark. Statistics on the reading comprehension materials are presented in Table \ref{tab:dataset-stats}.

%batches 1 & 2
\begin{table}[ht!]
\small
\begin{center}
%\resizebox{\columnwidth}{!}{\begin{tabular}{l|ll|ll}
\begin{tabular}{l|ll|ll}
\hline
 & \multicolumn{2}{c}{\bf OneStopQA}& \multicolumn{2}{|c}{\bf RACE} \\ \cline{2-5}
 &  Ele &  Adv &  Mid &  High\\ \hline
%Articles & 30 & 30 & 30 \\
%\# Paragraphs & 162 & 162 & 162 \\
 %Words & 11,789 & 14,718 & 1,657,122 & 7,362,796 \\
 Words per text & 112.3 & 138.6 & 232.1 & 354.1 \\
 \# Passages      & 162 & 162   & 6,409   & 18,728 \\
 \# Questions   & 486 & 486   &  25,421 & 62,445\\
\hline
\end{tabular}%}
\end{center}
\caption{\label{tab:dataset-stats} Statistics for OneStopQA and the RACE training set. The term \emph{passage} refers to a single paragraph in OneStopQA and an article in RACE.}
\end{table}

\subsection{OneStopQA Eye-Tracking Data}
\label{subsec:eyetracking-experiment}

We collected a dataset of eye movements for the 30 OneStopQA articles. The articles are divided into three 10-article batches with 54 paragraphs in each batch. Each participant read a single 10-article batch. Following the experimental setup of \cite{hahn2018}, a given batch is presented in one of two possible between subject conditions: \emph{Hunting} and \emph{Gathering}. In the Hunting condition participants are presented with the question prior to reading the text, while in the Gathering condition the question is provided only after the participant has completed reading the text. 

A single experiment trial consists of reading a paragraph and answering one reading comprehension question about it. In the Hunting condition, a trial has 5 pages in which the screen shows one page at a time. In the first page, the participant reads the question (henceforth \emph{question preview} page). In the second page, they read the paragraph. In the third page they read the question again. The fourth page retains the question, and also displays the four answers. After choosing one of the answers, the fifth page informs the participant on whether they answered the question correctly. The Gathering condition is identical to the Hunting condition, except that participants are not presented with the question preview page. Consequently, subjects in this condition have to be prepared for any question.

Each trial was randomly assigned to one of six conditions in a Latin square design, where each condition is a combination of one of the three questions and one of the two paragraph levels. The presentation order of the articles and the assignment of answers to A -- D letters was randomized. Eye movements were recorded using an EyeLink 1000 Plus eye tracker (SR Research) at a sampling rate of 1000Hz. The experiment duration was typically 1 - 1.5 hours. Further details on the eye-tracking experiment are provided in Appendix \ref{appendix-eyetracking}.

We collected data from 269 participants, with an average of 7.5 participants per trial (question - paragraph level pair). We excluded trials in which participants did not answer the question correctly, remaining with 6.3 participants per trial. The overall question answering accuracy rate  was 86.9\% in the Hunting condition and 81.9\% in the Gathering condition, which is lower ($p<10^{-4}$).\footnote{Satterhwaite's method on a mixed-effects model: $\text{correct}\sim\text{preview}+(\text{preview}||\text{subject})+(\text{preview}||\text{example})$.}

\begin{figure}[ht!]
\begin{center}
    \includegraphics[width=\columnwidth]{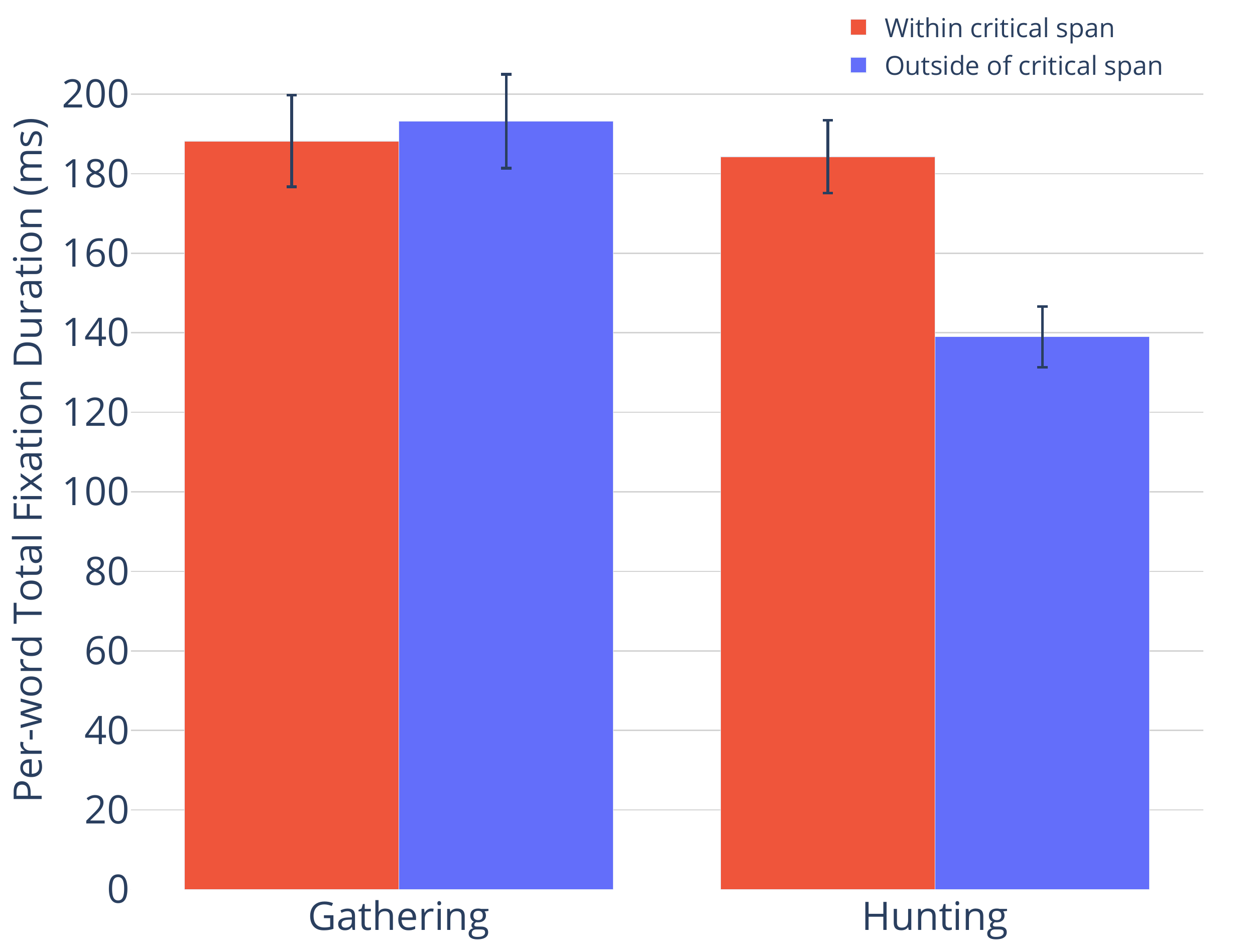}
    \caption{Mean Total Fixation Duration inside and outside the critical span in the Hunting (with question preview) and Gathering (without question preview) conditions. Error bars correspond to a 95\% confidence interval from a mixed-effects model that accounts for variation of fixation durations across subjects and questions. }
    \label{fig:span_reading_times}
\end{center}
\end{figure}

\section{Question Conditioned Gaze in Human Reading Comprehension}
\label{sec:reading-times}
\begin{figure*}[ht!]
    \centering
    \begin{subfigure}[b]{0.95\textwidth}
        \includegraphics[trim={1.5cm 2cm 0 0},clip,width=\textwidth]{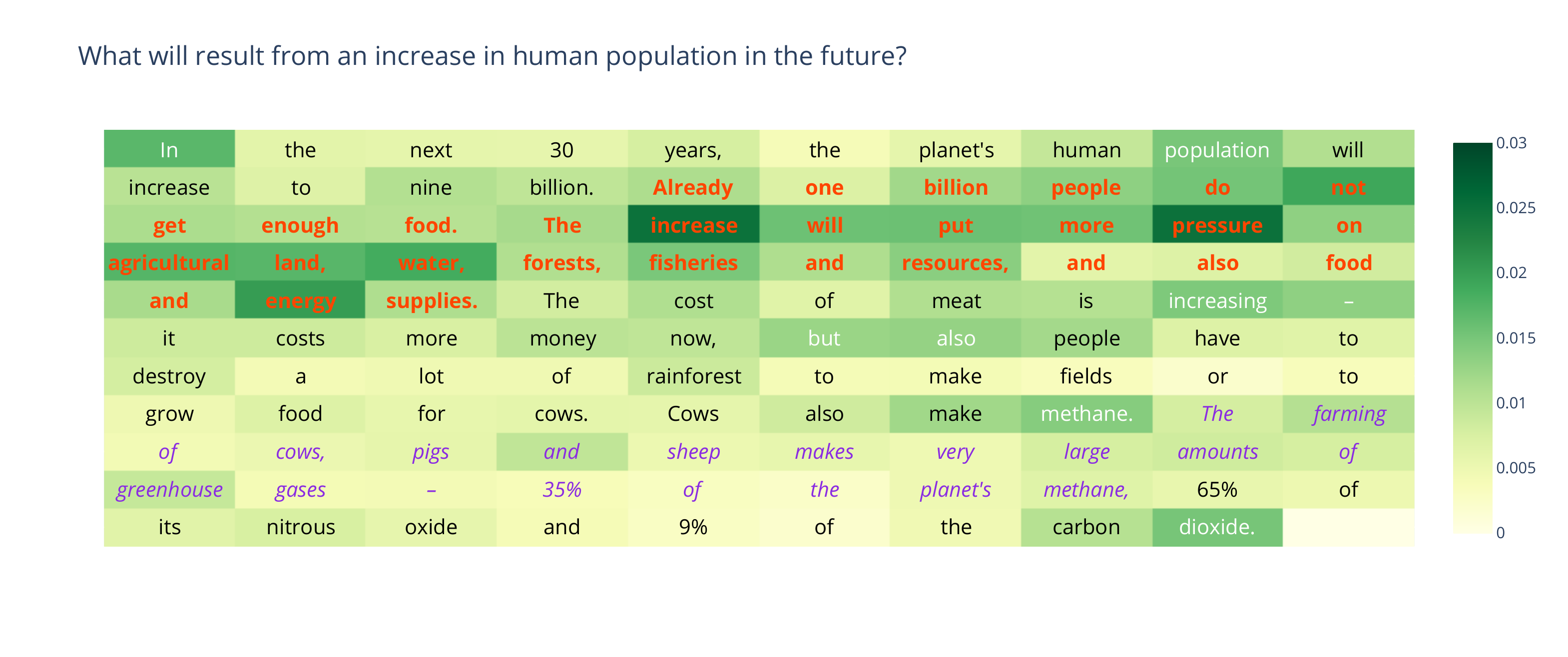}
        \caption{Hunting condition (with question preview)}
        \label{fig:gull}
    \end{subfigure}
      \par\medskip
    \begin{subfigure}[b]{0.95\textwidth}
        \includegraphics[trim={1.5cm 2cm 0 2.5cm},clip,width=\textwidth]{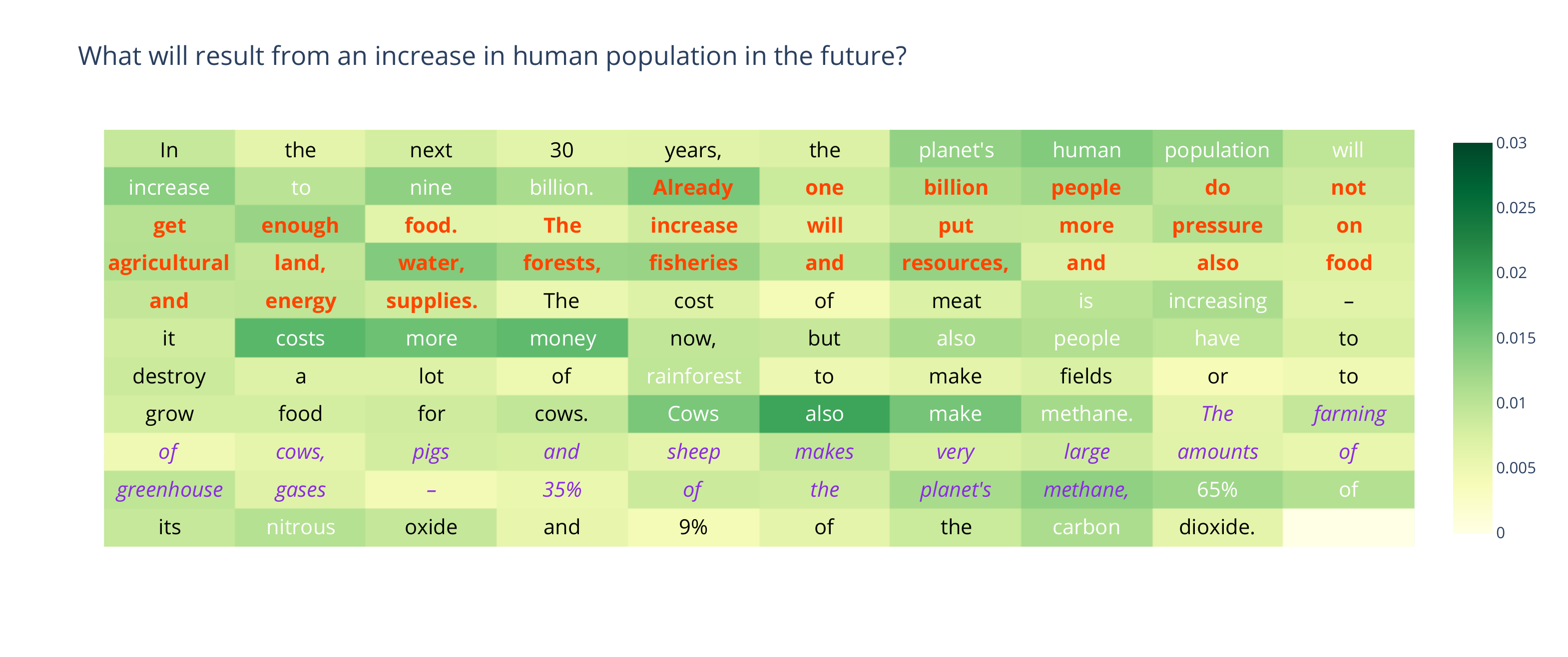}
        \caption{Gathering condition (without question preview)}
        \label{fig:tiger}
    \end{subfigure}
    \caption{Example of gaze distributions in the Hunting and Gathering conditions for an Elementary level paragraph. The color of each word corresponds to its Total Fixation Duration divided by the overall passage reading time, averaged across participants.  The critical span appears in bold red. The distractor span appears in purple italics.}
    \label{fig:examples}
\end{figure*}

We motivate our approach by demonstrating that human gaze distributions are strongly conditioned on the reading comprehension task. This conditioning has been previously established for the case of named entities \cite{hahn2018}, and we examine it here in a more general QA setting. Specifically, we consider speed-normalized Total Fixation Duration; for each subject, we take the Total Fixation Duration (i.e.\ sum of all the fixation times) on a word and normalize it by the subject's total reading time for the passage. Consider the example in Figure \ref{fig:examples}, where we visualize the speed-normalized gaze on each word averaged across subjects for the same question -- paragraph pair in the Hunting (with question preview) and Gathering (without question preview) conditions. As can be seen from the heatmaps, the gaze distributions are fundamentally different between these conditions. In particular, in the Hunting condition we observe a noticeable correspondence between gaze and the annotated critical span. Although the degree of correspondence between gaze and the critical span in the Hunting condition depends on the specifics of the question and the text, the presented example is representative of a large portion of our items.

To further substantiate this observation, in Figure \ref{fig:span_reading_times} we compare the average Total Fixation Duration within versus outside the critical span in both the Hunting and Gathering conditions. We observe that in the Hunting condition, reading times are significantly longer within the critical span compared to outside of the critical span ($p<10^{-15}$),\footnote{This and subsequent tests are calculated using Satterthwaite's method applied to a mixed-effects model that treats subjects and questions as crossed random effects. Using R formula notation, the model is $\text{gaze} \sim \text{span}*\text{condition} + (\text{span}||\text{subject}) + (\text{condition*span}||\text{example})$). Tests were performed with the lme4 and lmerTest R packages.} while in the Gathering condition they are slightly shorter within the critical span ($p<10^{-4}$). The difference between within-span vs outside-of-span reading times between Hunting and Gathering conditions is also significant ($p<10^{-15}$). We further note that the total reading time for the passage is shorter in the Hunting condition ($p<10^{-4}$), consistent with more targeted reading as compared to the Gathering condition.

While our analysis provides evidence for an increased concentration of gaze time around text that is critical for answering the question, the potential utility of human gaze is not limited to this aspect alone. Human gaze can be viewed as a soft form of text annotation that relates the entire text to cognitive load during processing. In particular, it can in principle provide valuable fine-grained information within the critical span.

\section{Method: Joint Question Answering and Human Gaze Prediction}

To test the effectiveness of utilizing human gaze data for enhancing the performance of a reading comprehension system, we trained a reading comprehension model to perform the same multiple choice task as the human subjects. We then conducted a series of controlled experiments to assess how the accuracy of the model is affected by providing it with human eye movements information.  

\subsection{Model}

\begin{figure*}
\begin{center}
    \includegraphics[width=0.78\textwidth]{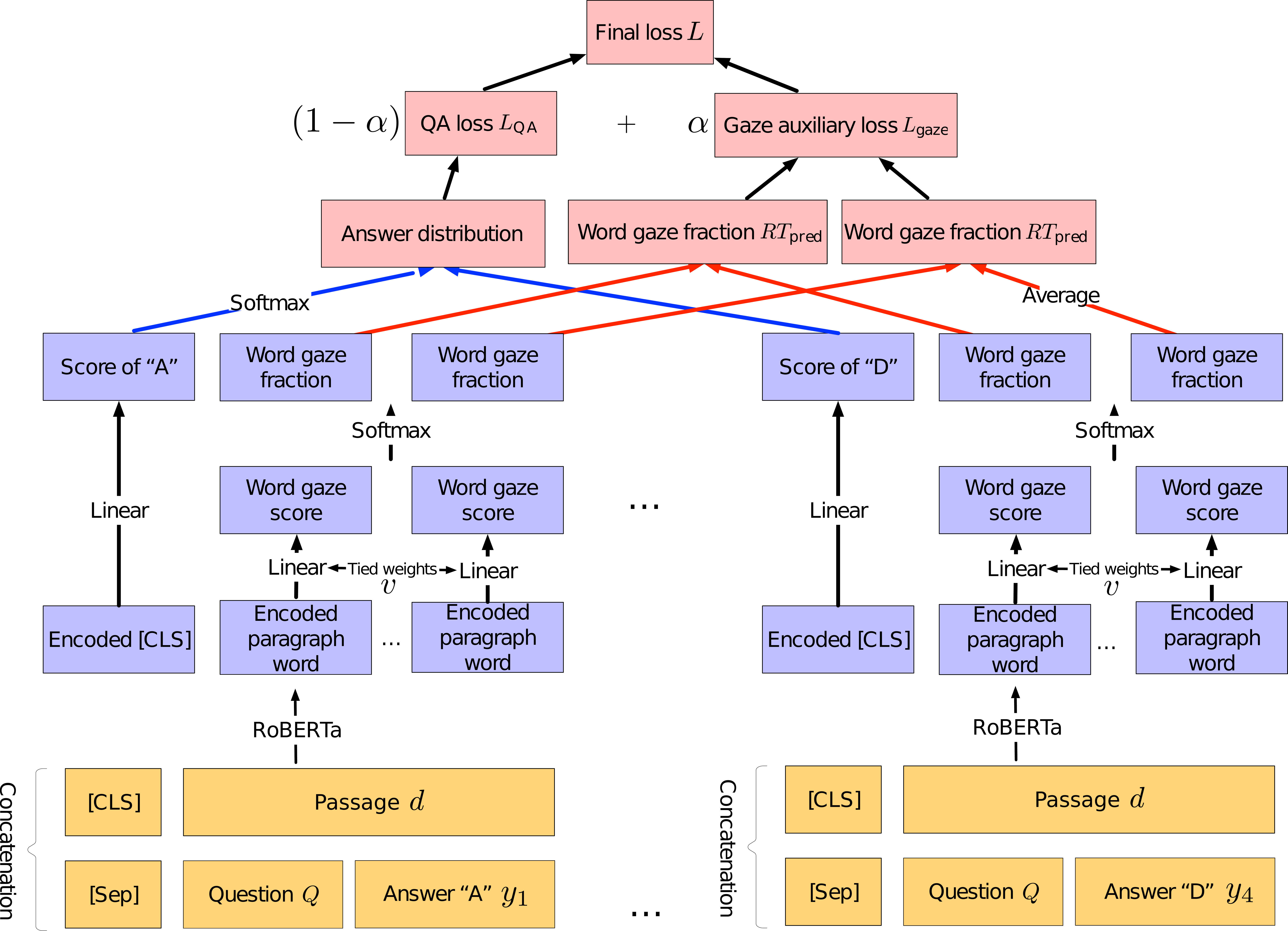}
    \caption{Model diagram. The model uses the standard transformer architecture for multiple choice QA, augmented to simultaneously predict human reading times over the passage.}
    \label{fig:model_diagram}
\end{center}
\end{figure*}

We utilize the RoBERTa transformer architecture, which has shown state-of-the-art performance on the multiple choice reading comprehension task \cite{roberta2019}. We experiment with both the Base and the Large variants of this model. To allow RoBERTa to benefit from the gaze data, we use multi-task learning with hard parameter sharing \cite{caruana1993}, and modify RoBERTa to jointly predict the answer to each question and the human gaze times allocated to each passage word. 

Each multiple-choice example is composed of the passage $d$, the question $Q$, and the four possible answers $\{y_{1},y_{2},y_{3},y_{4}\}$. We follow the standard procedure for using transformer architectures for multiple-choice tasks, concatenating the passage, question, and answer $[CLS,d,SEP,Q,y]$ for each possible answer $y$. The resulting string is encoded through RoBERTa. We then take the final embedding of the CLS token for each answer and run it through a logistic layer to return the probability of each answer being correct. This probability is used to calculate a cross-entropy QA loss term $L_{QA}$ for each example.
\begin{equation}
L_{QA} = -\log{p(y_{c})}
\end{equation}
Where $y_{c}$ is the correct answer for the question.

We additionally calculate an auxiliary loss based on gaze information. As in Figure \ref{fig:examples}, our reference metric $RT(w)$ is speed-normalized Total Fixation Duration ($TF$). Specifically, for each passage word $w$ and subject $s$, we consider the subject's Total Fixation Duration on the word $TF_{s}(w)$ normalized by the sum of all their fixation durations over the passage, and then average this quantity across all subjects who read the passage.
\begin{equation}
RT(w) = \frac{1}{S}\sum_{s}{\frac{TF_{s}(w)}{\sum_{w'}{TF_{s}(w')}}}
\end{equation}
In cases where RoBERTa's byte pair tokenizer \cite{sennrich2016} splits a single word into multiple tokens, we evenly split the gaze time associated with the word among the resulting tokens.

We take the encoding of each passage word at the last layer of RoBERTa for each candidate answer $y$ and add a linear layer parameterized by a weight vector $v \in R^d$ shared across all passage word positions, where $d$ is the RoBERTa embedding dimension. For each passage word $w$, this layer maps from the $d$-dimensional word embedding to a scalar gaze value. These values are put through a softmax layer, obtaining predictions $RT_{pred_{y}}(w)$ which are guaranteed to be between 0 and 1 for each word and sum to 1 for each passage, making them comparable to our normalized human gaze measurements $RT$. These predictions are then averaged across the four possible answers to obtain reading time predictions for each passage word $RT_{pred}(w)$. Finally, we compute the cross-entropy loss between the gaze predictions and observed human gaze.
\begin{equation}
L_{gaze} = -\sum_{w}{RT(w)\log{RT_{pred}(w)}}
\end{equation}

The final loss term is a convex combination of the gaze loss term and the answer prediction loss, where a hyperparameter $\alpha$ is the relative weight assigned to the gaze loss term: 
\begin{equation}
\label{loss_equation}
    L = (1-\alpha)L_{QA} + \alpha L_{gaze}
\end{equation}
Figure \ref{fig:model_diagram} presents a diagram of our model.

Our modelling approach is fundamentally behavioral, as it attempts to mimic human eye movements as an external behavior. It treats the model itself largely as a black-box, relying only on the model's final query-conditioned representations of the passage words. It is therefore also modular -- the RoBERTa model can be substituted with any QA model which provides passage word representations. Furthermore, our framework is compatible not only with the multiple choice variant of the QA task, but also with other answer output formats.

We also note that the standard multiple choice QA transformer architecture requires a copy of the passage and the question for each answer, and thus the reading time predictions are generated for each copy and averaged. In QA models were the query and passage are encoded only once, such averaging would not be required. Further, other architectures are conceivable for joint multiple choice QA and gaze prediction. In particular, one may consider architectures which do not include the answers for gaze prediction; for example, through soft parameter-sharing multi-task approaches. We chose hard parameter sharing as it enables predicting gaze with only a minimal architecture change and a small number of additional parameters to the standard multiple choice QA transformer model. 

\subsection{Training Procedure}

Each experiment consists of a training set of QA examples from OneStopQA accompanied by gaze data, a development set, and a test set. For each experiment, we fine-tune an initial model for 15 epochs for each $\alpha \in [0, .2, .4, .6, .8, 1.0]$. We pick the epoch and $\alpha$ that have the highest question-answering accuracy on the development set and report accuracy on the test set. For experiments on OneStopQA, we perform five-fold cross validation where each fold has 18 training articles, 6 development articles and 6 test articles. Each article appears three times in train, once in dev and once in test across the 5 folds.

\subsection{Conditions}

We test two initial models:

\begin{enumerate}
\setlength\itemsep{0.2em}
    \item \textbf{No RACE fine-tuning} using RoBERTa that has not been fine-tuned for QA on RACE. This experiment shows the value of incorporating eye-tracking data in data-scarce scenarios where only a small amount of data is available for fine-tuning on the given task.
    \item \textbf{With RACE fine-tuning} using RoBERTa that has been fine-tuned on RACE to perform multiple choice question answering, following the procedure in \cite{roberta2019}.
\end{enumerate}
    
For each fine-tuning regime, we test the model for two levels of generalization:

\begin{enumerate}
\setlength\itemsep{0.2em}
    \item \textbf{Within-domain} where we use our five-fold cross validation setup to train and test on OneStopQA.
    \item \textbf{Out-of-domain} where we train on all 30 OneStopQA articles and use the RACE dev and test sets for development and testing. 
\end{enumerate}

We note that in addition to the quality assurance issues with RACE mentioned in Section \ref{rc_data}, the out-of-domain RACE evaluations are particularly challenging due to substantial differences in the genres and questions types between OneStopQA and RACE, and the small size of OneStopQA as compared to RACE.

\begin{table*}[ht!]
\begin{center}
\begin{tabular}{cl|c|c|c|c|c|c|c|c} 
&      \multicolumn{1}{c}{}& \multicolumn{4}{c}{\bf OneStopQA} & \multicolumn{4}{c}{\bf RACE}\\ \cline{3-10}
&                     & \multicolumn{2}{c|}{No RACE} & \multicolumn{2}{c|}{With RACE}  & \multicolumn{2}{c|}{No RACE} & \multicolumn{2}{c}{With RACE}  \\ 
&                     & \multicolumn{2}{c|}{Fine-tuning} & \multicolumn{2}{c|}{Fine-tuning} & \multicolumn{2}{c|}{Fine-tuning} & \multicolumn{2}{c}{Fine-tuning} \\ \cline{3-10}
&  & Base & Large & Base & Large& Base & Large& Base & Large  \\ \hline

&No OneStopQA Fine-tuning  & 23.5 & 23.1 & 68.5 & 85.8 & 24.2 & 24.3 & 68.4 &82.9  \\ 
&With OneStopQA Fine-tuning & 48.9 & 62.9 & 68.3 & 85.9 & 36.8 & 42.9 & 68.4 & 82.9\\ \hline
\parbox[t]{2mm}{\multirow{5}{*}{\rotatebox[origin=c]{90}{+ Aux. Loss}}} & Question--Passage Similarity &54.7 & 78.9 & 72.3 & 87.3 &\bf 41.2 & 52.6 & 68.2& 82.9\\
&Gaze Gathering Dundee & 55.9 & 78.2 & 72.5 & 86.6 & 39.3 & 52.5 & 68.4 &  82.9\\
&Gaze Gathering OneStopQA & 54.7 & 80.2 & 71.5 & 87.0 & 39.6 & 48.4 & 68.4 & \bf 83.0\\
&Critical Spans OneStopQA & 54.7 & \bf 80.7 & 70.1 & 86.5 & 40.4 & 51.6 & 68.4 & 82.9 \\
\cline{2-10}
&Gaze Hunting OneStopQA & \bf57.1 &  80.5 & \bf 73.1 & \bf 88.0 & 41.1 &\bf 53.0 & \bf 68.5 & \bf 83.0\\
\hline
\end{tabular}
\end{center}
\caption{Question Answering accuracy for RoBERTa Base and Large on OneStopQA and RACE. RACE Fine-tuning denotes whether the model has been first fine-tuned for QA on RACE. %Base and Large correspond to RoBERTa Base and RoBERTa Large. 
The first two rows are baselines without an auxiliary loss, without and with QA fine-tuning on OneStopQA. The following four rows are baselines fine-tuned for QA on OneStopQA with auxiliary loss, using four different alternatives for measuring the importance of each passage word. The last row is our primary model variant, which uses gaze in the Hunting condition. All the results with OneStopQA fine-tunings are averaged over three runs of the model.}
\label{table:results}
\end{table*}

\subsection{Baselines}

We compare our model with two baselines which do not use auxiliary loss. We further introduce four auxiliary loss models, which replace Hunting condition gaze with alternative information sources for measuring the importance of each passage word.

\subsubsection*{No Auxiliary Loss} 
These two baselines do not utilize the auxiliary loss during model fine-tuning.
    \begin{enumerate}
    \setlength\itemsep{0em}
        \item \textbf{No OneStopQA Fine-tuning} The model is not fine-tuned for QA on OneStopQA.
        \item \textbf{With OneStopQA Fine-tuning} The model is fine-tuned for QA on OneStopQA.
    \end{enumerate}
    
\subsubsection*{With Auxiliary Loss}
These four models are fine-tuned for QA on OneStopQA, and use an auxiliary loss where gaze in the Hunting condition is replaced with other ways for weighting each word in the passage. 
\begin{enumerate}
\setlength\itemsep{0em}
    \item \textbf{Question--Passage Similarity} In this baseline, the auxiliary information is based on the similarity between the question and each passage word. We encode the question and the passage separately with an off-the-shelf encoder (here, RoBERTa that has not been fine-tuned for question-answering) and compute the dot-product between each encoded passage word and the final encoding of the question's CLS token. These values are then normalized by applying a softmax function.
    \item \textbf{Gaze Gathering Dundee} Here, we utilize gaze data from the Dundee corpus \cite{dundee}, allowing us to examine the benefit of predicting gaze on the same texts used for QA, versus unrelated texts. We split each Dundee article into passages of size equal to the average OneStopQA passage (125 words), yielding 453 passages. We then normalize the average Total Fixation Duration across Dundee's 10 subjects as for OneStopQA. In each training step, we predict answers on one batch of OneStopQA questions and gaze distributions on one batch of Dundee paragraphs chosen at random, and perform a step of gradient descent. This interleaved procedure is similar to that used by Barrett et al. \shortcite{barrett2018}, and is analogous to the other baselines, where we predict answers on one batch of OneStopQA examples and gaze distribution on those same examples for each gradient descent step. 
    \item \textbf{Gaze Gathering OneStopQA} In this method, we use gaze data from the Gathering variant of the OneStopQA reading experiment where subjects do not see the question before seeing the paragraph they will later be questioned about, and hence their gaze is necessarily not question-dependent.
    \item \textbf{Critical Span Annotations OneStopQA} In OneStopQA, each question includes a manual annotation which indicates the span in the passage which is critical for answering the question. We assign a gaze value of $1$ to the tokens within the span and $0$ to those outside it, and normalize with softmax as before. This corresponds to a theoretical subject who looks equally at each word within the critical span and not anywhere else in the passage.
\end{enumerate}
We note that the last two baselines are new methods for improving machine QA using human-generated behavioral data (gaze and span annotations) that have not been previously proposed in the literature, and constitute very strong alternatives to our model.

\section{Experimental Results}

Our results are summarized in Table \ref{table:results}. All the results involving OneStopQA fine-tunings are averaged across three runs. In the following, p values are indicated when the difference in the performance of the compared models is statistically significant at the $p<0.05$ level.

Fine-tuning the model for QA on OneStopQA is most beneficial in the two resource-lean regimes when the model has not been previously fine-tuned on RACE ($p<10^{-10}$, Wald test). Similarly, adding auxiliary loss to the QA model in these two regimes has a substantially larger impact on model performance compared to performing prior fine-tuning on RACE ($p<10^{-8}$ for all baselines).

In our within-domain evaluations on OneStopQA, we observe improvements of our model over all the baselines in all evaluations, except for the case of the Large model without RACE fine-tuning where our model comes second.
We also observe improvements in the out-of-domain evaluations on RACE. When the Large model is fine-tuned for QA only on OneStopQA, it obtains an accuracy of $53.0$, reflecting a $0.4$ improvement over the strongest baseline. The Base model comes second in this evaluation. When first fine-tuning the model for QA on RACE, then performing additional fine-tuning on OneStopQA, the Base model obtains an improvement of $0.1$ over the strongest auxiliary loss baseline. For the Large model we observe a similar improvement when using gaze, with the same performance in the Hunting and Gathering conditions.

Interestingly, we do not observe a consistent ordering in the performance of the baselines. In particular, we do not observe a clear advantage of using gaze in the Gathering condition over Question--Passage Similarity. We also obtain comparable performance when gaze data in the Gathering condition comes from OneStopQA and Dundee. Notably, in nearly all the evaluations our model performs better compared to the manual Critical Span Annotation baseline. We hypothesize that this may be because the annotated spans do not capture potential inter-annotator variation in span annotations, as well as within-span information which is informative for our task.

We note that while the gains over the strongest baselines are not statistically significant at the $.05$ level, the overall consistent pattern across evaluation regimes suggests the promise of using Hunting gaze data as the target of the auxiliary loss objective over any other single baseline. 
Finally, we note that an $\alpha$ of $0.2$ -- $0.4$ was most often chosen.

\section{Conclusion}
\label{sec:discussion}

We present a framework for performing automated reading comprehension in a human-like fashion, yielding performance gains for a state-of-the art reading comprehension model. Our work also contributes to the study of human reading, providing evidence for a systematic conditioning of human reading on the reading comprehension task. 
In the future we intend to study the relation between gaze and internal model attention, and further explore the relation between gaze, task and task performance in QA and well as other tasks.

\section*{Acknowledgments}
We gratefully acknowledge support from Elemental Cognition and from NSF grant IIS1815529, a Google Faculty Research Award, and a Newton Brain Science Award to RPL.

\bibliographystyle{acl_natbib}
\bibliography{emnlp2020}

\appendix
\section{Supplemental Material: OneStopQA Eye-Tracking Experiment}
\label{appendix-eyetracking}

\subsection*{Eye Tracker}
We used a Tower Mount Eyelink 1000 Plus eye tracker (SR Research) at a sampling rate of 1000Hz. Eye movements were recorded for participants' dominant eye.

\subsection*{Monitor}
The experiment was presented on a 27inch monitor (Dell U2715H) with a display area of 597mm$\times$336mm, resolution of 2560px$\times$1440px and refresh rate of 60Hz. Participants' eye level was 750mm away from the top of the monitor's display area and 795 away from its bottom. In this setup participants eyes were about 45mm below top of the monitor's display, approximately at the same height as the top most position of the text.

\subsection*{Controller}
Participants used a controller (Logitech Gamepad F310) during the experiment. The button A was used for proceeding to the next page after finishing reading as well as for confirming the answer selection. The four buttons of the directional pad were used for choosing answers.

\subsection*{Text}
We used the Lucida Sans Typewriter monospace font, with font size of 25pt (each letter occupying 19px$\times$38px). We used triple spacing (76px) between lines. The top left position of the questions and the paragraphs was (300, 186) with a text area width of 1824px (96 characters). Questions were 1-2 lines and paragraphs were 3-10 lines. Answers were presented in a cross arrangement, with text width of 700px, and were 1-3 lines.

\subsection*{Calibration}
We used 9 point calibration with bulls-eye targets (18px outer circle 6px inner circle). Calibration was performed at least 3 times during the experiment: once at the beginning of the experiment and once after each of two breaks. Calibration was also performed upon failure to trigger the text at the beginning of a trial as described below. The experimenters were instructed repeat calibration until an average validation error below \ang{0.3} was reached. 

\subsection*{Text Triggering and Recalibration}
Prior to the presentation of the question preview, paragraph and question, participants were presented with a page presenting a fixation target located at (300, 186), the same position as the first letter of text on the following page. The targets were \textbf{q} for the question preview, \textbf{p} for the paragraph and \textbf{Q} for the question. The presentation of the following text page was triggered by a fixation of at least 250ms within a 39px$\times$48px rectangular area centered around the 19px$\times$38px area of the target letter. This corresponds to a horizontal margin of about half a letter width, and vertical margin of about quarter of a line space around the target letter. 

Failure to produce a 250ms fixation within 4 seconds on the first target of the trial (\textbf{q} target in the Hunting condition and \textbf{p} target in the Gathering condition), automatically triggered recalibration. For subsequent trial targets (\textbf{p} and \textbf{Q} in the Hunting condition and \textbf{Q} in the Gathering condition) the next page was presented even if the participant was not able to produce a 250ms fixation on the target letter within 4 seconds.

\end{document}